\documentclass[preprint]{elsarticle}
\usepackage[draft]{todonotes}
\usepackage{array}
\usepackage{wrapfig}
\usepackage{booktabs}
\usepackage{graphicx}
\usepackage{adjustbox}
\usepackage{caption}
\usepackage{subcaption}
\usepackage{amsmath}
\usepackage{hyperref}
\usepackage{listings}
\graphicspath{{images/}}

\def\BibTeX{{\rm B\kern-.05em{\sc i\kern-.025em b}\kern-.08emT\kern-.1667em\lower.7ex\hbox{E}\kern-.125emX}}

\definecolor{driesColor}{RGB}{200,0,150}

\begin{document}

\begin{frontmatter}
\title{An Automated Engineering Assistant:\\Learning Parsers for Technical Drawings}

\author[add1]{Dries Van Daele}
\ead{dries.vandaele@cs.kuleuven.be}

\author[add2]{Nicholas Decleyre}
\author[add2]{Herman Dubois}
\author[add1]{Wannes Meert}

\address[add1]{KU Leuven, Dept. of Computer Science, Leuven, Belgium}
\address[add2]{Saint-Gobain Mobility $\mid$ Engineered Components (Seals), Kontich, Belgium}

\begin{abstract}
From a set of technical drawings and expert knowledge, we automatically learn a parser to interpret such a drawing. This enables automatic reasoning and learning on top of a large database of technical drawings. In this work, we develop a similarity based search algorithm to help engineers and designers find or complete designs more easily and flexibly.
This is part of an ongoing effort to build an automated engineering assistant.
The proposed methods make use of both neural methods to learn to interpret images, and symbolic methods to learn to interpret the structure in the technical drawing and incorporate expert knowledge.
\end{abstract}

\end{frontmatter}
\section{Introduction}

\begin{figure}[bt]
  \includegraphics[width=\linewidth]{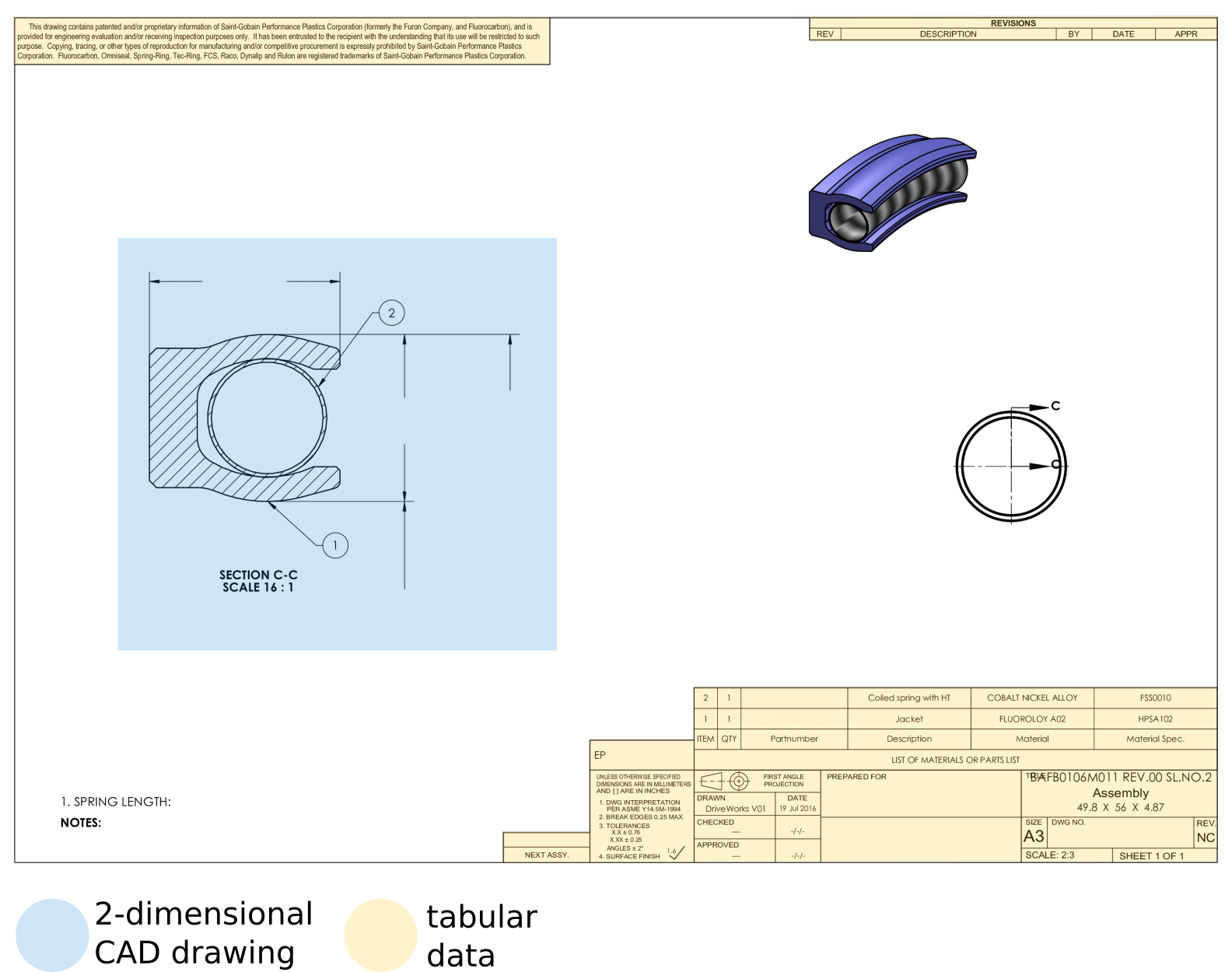}
  \caption{A technical drawing with highlights indicating the 2D CAD drawing and the tabular data}
  \label{fig:technical_drawing}
\end{figure}

Technical drawings are the main method in engineering to visually communicate how a new machine or component functions or is constructed. 
They are the result of a design process starting from a set of specifications that the final product needs to comply with. This design process follows a number of strict and soft rules (e.g., material choice as a function of temperature).
Figure \ref{fig:technical_drawing} shows a typical example containing both a 2D and 3D visualisation of the object, and a material list in tabular form specifying its parts and properties. They are carefully crafted documents that act as key deliverables at the end of a design process. As such, they contain a wealth of information. Furthermore, information is laid out according to generally applied conventions.

Engineering companies have a large database of previous designs, potentially going back decades. They are often underutilized because previous designs can only be searched for by title or by using a limited set of textual annotations. Ideally, however, this database can also be used to: (1) given a technical drawing, find other relevant drawings in a large database of previous designs; (2) given a partial description, find designs that would complete the partial design. In this work we present an approach that can extract the knowledge in a technical drawing and thus improve the search capabilities significantly to achieve the aforementioned tasks and assist engineers during the design process.

To be able to use the data encapsulated in technical drawings, we need
to parse the information they contain, both tabular as well as visual,
and translate this to a representation that can be handled by automated
systems. Furthermore, such a system should be able to deal with both
recent digital drawings and historical analog drawings. The latter is
important because a great amount of information is captured in legacy
drawings. Ideally, extracting the information can be done using a parser, which is a
small computer program. The main challenge is that writing and maintaining such a parser is a time-consuming and expensive
task. Furthermore, it is error prone since this requires an expert to
explain subtle rules to an analyst or a programmer. The approach we
present here will learn such parsers directly from expert feedback on
the original drawing and allow its output to be used in automated
tasks such as searching relevant designs.

Learning a parser requires expert input. Here, this input takes
the form of annotated technical drawings. Providing such annotations
is a trivial task for domain experts. The required number of drawings
that need to be annotated is mainly dependent on the number of
variations or templates that need to be recognized. Fortunately, since all technical
drawings within an organisation are expected to be (loosely) based on
a limited set of templates, the number of drawings that need to be
annotated is also limited.

The approach presented in this work is a hybrid approach that
utilizes both neural methods and reasoning-based methods. This
combination is necessary to capture the full range of
information. Neural methods such as deep convolutional neural networks
are used because they are the state-of-the-art in
image recognition algorithms. Despite their success, however, automatic
analysis and processing of engineering drawings is still far from
being complete \citep{moreno2018new}. This is in part due to the demand of neural methods for large amounts of training data. Such data is not always available, even though an expert might be capable of summarizing part of the knowledge or the parser in just a few abstract concepts. To exploit
this expert knowledge, we also utilize reasoning based methods such as
inductive logic programming and statistical relational learning. Such
a hybrid approach that combines methods that are data-driven with
methods that are knowledge-driven is gaining in popularity since real-world tasks tend to require Hybrid AI where human knowledge or reasoning is directly integrated \citep{Manhaeve2018NIPS,mao2018the}.

In this work we explain five contributions.
First, we introduce the use of ILP to learn parsers from data and expert knowledge to interpret technical drawings.
Second, we introduce a novel bootstrapping learning strategy for ILP.
Third, we introduce a deep learning architecture that learns a meaningful summarization of CAD drawings.
Fourth, we introduce a similarity measure to find related technical drawings in a large database.
Finally, the efficacy of this method is demonstrated in a number of experiments on a real-world data set.

\section{Overall system}
\label{sec:methodology-overview}
\begin{figure}
  \includegraphics[angle=270,width=\linewidth]{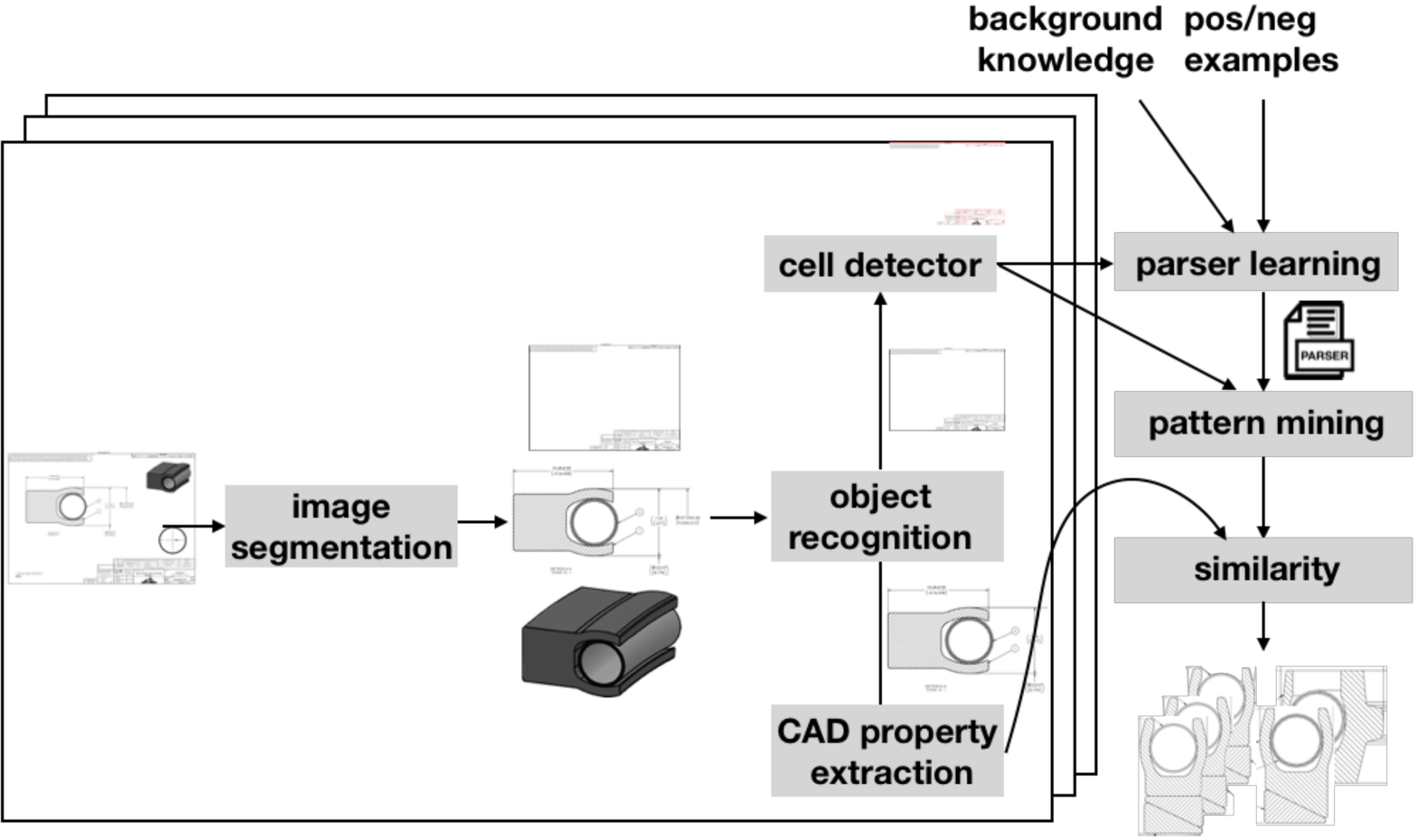}
  \caption{Overview of the technical drawing similarity proposal
    system.}
  \label{fig:overview}
\end{figure}

The goal of extracting the key information contained in a technical drawing (e.g., Figure \ref{fig:technical_drawing}) is achieved by focusing our attention on two central components: (1) the tabular information with a list of parts and materials, author, date, etc.; and (2) the two-dimensional CAD-drawing visualizing the precise component shape and dimensions.

The modular structure of our implementation is shown in Figure
\ref{fig:overview}. First, the central components of the drawing are
extracted using image segmentation, and are identified by the object
recognition module (Section \ref{sec:preprocessing}). Then, we use the
learned parser to extract properties from the tabular data (Section
\ref{sec:ilp-and-srl}) and the CAD-drawing (Section
\ref{sec:CNN}). Finally, the output of these processing steps is
synthesized into a feature vector on which we impose a similarity 
measure, over all possible designs (Section \ref{sec:similarity-measure}).


We had access to 5000 technical drawings, but used only the finished drawings. About $1\%$ of the drawings were unfinished (e.g. flagged as in progress, overlapping objects) and removed.

\section{Identifying technical drawing elements}
\label{sec:preprocessing}
Archived technical drawings are digitized to varying degrees. Because of this, we consider as a baseline the case where the technical drawing is represented as a bitmap image.

\subsection{Image segmentation}

Extracting all the relevant elements contained in the image (e.g.,
tables, drawings) is achieved using conventional computer vision
methods. The image is partitioned into its main segments using DBSCAN with $\epsilon=30$ and minimum
  points set to 0.001\% of total pixels, thus $\approx 75$ points \citep{Ester:1996:DAD:3001460.3001507}. Since a technical
drawing employs white space 
to distinguish central layout elements, such a density-based method is highly effective at segmenting images into their constituent elements. No errors were observed in the segmentation of the drawings.

\subsection{Object recognition}
\label{sec:object-recognition}
Next, the system recognizes what each image segment represents by classifying them as one of three possible classes: `tables', `two-dimensional CAD drawings', and `irrelevant' segments. This is a rather simple task as these classes are highly visually distinctive. High predictive accuracy is achieved with a small CNN classifier. For our implementation we relied on the PyTorch library \citep{paszke2017automatic}. This classifier was constructed with three convolution layers and three fully connected layers. It was trained against 318 technical drawings that were annotated by an expert, for a total of $3000$ image segments ($318$ tables, $372$ CAD drawings, $2310$ irrelevant segments). No classification errors were made on a test set of 53 technical drawings containing 500 segments.

In case a table is recognized we apply one additional operation to identify the cells by applying a contour detection algorithm \citep{SUZUKI198532} provided by the OpenCV library. The cells of each table are subsequently processed by the \emph{parser learning}-module, which is specifically aimed at learning how to read the tables contained in the technical drawings referenced throughout this paper, such as a material bill (see Section~\ref{sec:ilp-and-srl}).
In case a 2-dimensional CAD drawing is recognized, the image data is passed on to the \emph{CAD property extraction}-module that has learned how to recognize the application dependent parts of an image (see Section~\ref{sec:CNN}).

\section{Extracting properties from tables} 
\label{sec:ilp-and-srl}
The data contained in a technical drawing is laid out in a manner that
facilitates human interpretation. Tabular data in particular tends to
be organised both spatially and through explicit annotation. Common
examples of spatial structuring involve assigning related cells to
common rows or columns, while assigning unrelated cells to different
subtables or distant cells. Particularly useful are cells that contain
unambiguous keywords such as attribute names. These are helpful to gain insight in the structure of a table. They serve as anchors to cells that are less distinctive on their own but can easily be described relative to them. 

The application at hand does not only require us to parse a table, but
also demands that we learn how to interpret its spatial
organisation. A small computer program is required to parse these
custom drawings. Programming a parser for each type of drawing is not
only an expensive and time consuming task to build and maintain, but
also error-prone. Various errors are potentially introduced while
programming a parser. First, the structure of such technical drawings
needs to be explained to a non-expert, i.e. a programmer, who
interprets the instructions. Second, the tables are typically not
simple rectangular tables. They thus require a non-trivial parser that
is difficult to understand. Third, a design can deviate slightly or
change over time requiring periodic maintenance and potentially
leading to software erosion. Ideally these programs would be derived
directly from the expert's knowledge, and be easily updated when new
designs appear. This is possible by means of machine learning
techniques that learn programs from examples. The examples in this
setting are obtained by annotating technical drawings, a task that is
trivial for a domain expert.

The highly relational nature of tabular data and the ease with which
tables can be sensibly navigated by visiting adjacent cells suggests
the use of Inductive Logic Programming (ILP). ILP systems are
particularly suitable for learning small programs from a limited amount of complex input
data.  When learning the programs covered in this work using ILP, we benefitted in particular
from the ability to learn recursive definitions (e.g., row $n$ is
defined by row $n+1$) and reuse learned target labels (e.g., first
learning what a header row is helps to define what a content row is).
A logic program consists of a set of definite clauses. A definite clause can be interpreted
as a rule. One such definite clause might look like:
$h(a, X) :- b_1(a, X), b_2(X).$
where $h(a, X)$, $b_1(a,X)$, and $b_2(X)$ are literals whose arguments
can either be constants or logical variables. Constants are
denoted using lowercase letters or numbers, while variables are
uppercase letters. Disjunction is represented using `;' and conjunction
using `,'. 
Programs obtained through ILP can be augmented with probabilistic
distributions. This combination is referred to as Statistical
Relational Learning \citep{Getoor:2007:ISR:1296231,
  DeRaedt:2010:LRL:1952055} and can naturally deal with uncertain
input information. This is a useful property, because the input data for
the parser is provided by other machine learning models. For example, the OCR
engine returns a distribution over characters instead of a 100\% certain
prediction.

\subsection{Inductive Logic Programs for Parsing}
\label{sec:ILP-set-up}
An inductive logic programming system learns from relational data a
set of definite clauses. Given background knowledge $B$, positive
examples $E^{+}$ and negative examples $E^{-}$, it attempts to
construct a program H consisting of definite clauses such that $B
\wedge H$ entail all, or as many as possible, examples in $E^{+}$, and
none, or as few as possible, of those in $E^{-}$.

We thus need to supply three types of inputs.

First, a set of training data, examples $E$, that contains the
properties to describe a cell in a technical drawing. An example can
be:%

{\setlength{\itemsep}{0pt}\setlength{\topsep}{0pt}\setlength{\partopsep}{0pt}\setlength{\parsep}{0pt}\setlength{\parskip}{0pt}%
\begin{itemize}
\item[--] \emph{Cell text:} The textual contents of each cell. Tesseract 4.0 is used to recognize cell contents \citep{Smith:2007:OTO:1304596.1304846}.
\item[--] \emph{Cell location:} The cell's bounding box information (i.e.\@ (x,y) coordinates and cell width and height).
\end{itemize}
}

Second, a label for each cell (e.g., author, bill of materials,
quantity). A cell can be annotated with multiple labels (e.g., a cell
can be a quantity in the bill of materials). Depending on which target
label we want to learn, we split the set of examples $E$ in a tuple
$(E^{+}, E^{-})$ where $E^{+}$ contains the examples associated with a
cell that has the target label and $E^{-}$ those examples that do
not. For standard ILP, the learning task is defined for one target
label, so we repeat the standard ILP task for each label in the set of
labels.

Third, we can provide background knowledge $B$ that contains
generally applicable knowledge for the problem at hand and remains
unchanged across examples. In this case we provide:

{\setlength{\itemsep}{0pt}\setlength{\topsep}{0pt}\setlength{\partopsep}{0pt}\setlength{\parsep}{0pt}\setlength{\parskip}{0pt}%
\begin{itemize}
\item[--] \emph{Relative cell positions.} Relations capturing which cells are adjacent to each other, and in which direction (horizontally or vertically) based on their bounding boxes.
\item[--] \emph{Numerical order.} The successor relationship. Although not essential, it is useful for learning concise, recursive rules.
\end{itemize}
}

The output of ILP, the program $H$, is a set of definite clauses of
the form `\texttt{author(A) :- cell\_contains(A, drawn)}' which can be
read as the rule: Cell A contains the author if it contains the word `drawn'.


\subsection{ILP with bootstrapping}
It is expected that learning programs to properly parse the target labels in P will prove simple for some targets and more challenging for others. We propose a bootstrapping extension that supports the construction of sophisticated programs by allowing them to employ the simpler ones in their definition. This is loosely inspired by the ideas raised in \cite{Dechter:2013:BLV:2540128.2540316}, but applied to the ILP setting.

This corresponds to a variation of the previously discussed ILP set-up
where a dependency graph $G$ is used. The nodes in this directed
acyclic graph each represent a possible target label and the edges
represent dependencies between those labels. A dependency indicates
that one target label might have a natural description in function of
another. Although we allow for this dependency graph to be specified
manually, our method defaults to a fully automated approach where
standard ILP is first applied to learn programs for each target. Then
targets are ranked according to, first, ascending $F_{1}$ score on the
training data and, second, the size of the program in number of
literals. Each target in the ranking then has all subsequent targets as
its dependencies. Finally, ILP with bootstrapping learns targets in
the order specified by a correct evaluation order of $G$, and extends
the background knowledge $B$ for each target with the programs
constructed to parse its dependent target labels. When learning
program H using bootstrapping to capture a particular target label l,
we define its extended background knowledge $B' = B \wedge
(\bigwedge_{i \in descendants(G, l)} H_i)$, where $H_i$ is the program
trained for target label i.

\begin{figure}[ht]
  \centering
  \begin{subfigure}{\linewidth}
    \centering\includegraphics[width=\linewidth]{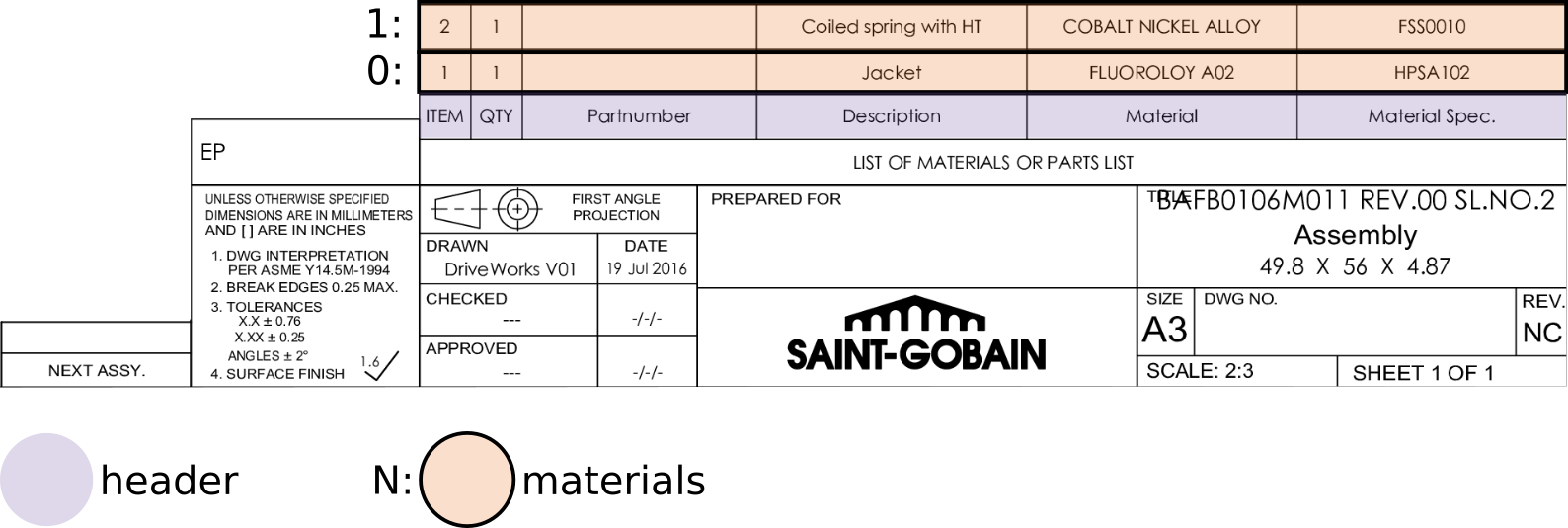}
    \subcaption{\label{fig:fig1} A table excerpt from a technical drawing. Its header and materials are highlighted.}
  \end{subfigure}%
  \\
  \begin{subfigure}{\linewidth}
  \begin{minipage}{0.45\linewidth}
    \begin{lstlisting}[basicstyle=\footnotesize]
% Materials hypothesis
materials(A,B) :-
   zero(A),
   above_below(B,C),
   header(C).
   \end{lstlisting}
  \end{minipage}
  \begin{minipage}{0.45\linewidth}
    \begin{lstlisting}[basicstyle=\footnotesize]
      
materials(A,B) :-
   succ(C,A),
   above_below(B,D),
   materials(C,D).
\end{lstlisting}
\end{minipage}
\\
\begin{lstlisting}[basicstyle=\footnotesize]
% Header hypothesis
header(A) :-
   above_below(A,B),
   cell_contains(B, `LIST').
\end{lstlisting}
 
\subcaption{\label{fig:fig2} header/1 covers any cell located directly
  above a cell containing the word `LIST'. materials/2 parses the
  indexed parts of the materials table. Its first argument is the
  index and its second argument represents the cell. materials/2
  consists of two clauses. The first clause anchors the table by
  considering row 0 to consist of the cells above the header. It
  employs header/1 in its definition. The second, recursive clause
  indicates that the index is incremented whenever a row is located
  above another.}

  \end{subfigure}
  \caption{Figure \subref{fig:fig1} provides an illustration of the materials table and its header. Listing \subref{fig:fig2} shows the associated program learned using bootstrapping.}
  \label{fig:materials/2}
\end{figure}

\subsection{Experiment set-up}
\label{sec:ILP-results}
The ILP system Aleph \citep{srinivasan2001aleph} is used to learn possibly recursive programs that parse the chosen targets from the tabular data, ranging from the document's author and its approval date to the attributes covered in the materials table and its indexed components.

Training data consists of a set of fully labeled technical drawings. A custom data labeling tool with a web-based graphical interface was constructed to support domain experts in labeling drawings.

Using this tool, 30 technical drawings with on average 50 cells were labeled with 14 different labels. For each target label, examples that contain that label form its positive example set, while negative examples are automatically derived by taking the complement of all possible examples for that target with its positive example set.

The labeled data is split in a training set consisting of 10 drawings,
and a test set containing the remaining 20. Since the choice of
training data can heavily affect the capability for finding rules that
properly generalize, experiments are repeated 5 times on random
samples of the training data. Because the order in which training
examples are presented can also affect the rules identified by the
coverage-based algorithm employed by Aleph, repeat experiments are
performed even when all training data is available for learning, as a
sample then corresponds to a different order in which the examples are
presented to the learner.

When inducing programs using Aleph, we employ a proof depth of 12, a clause
length of 5, and an upper bound of 60 000 on the number of nodes that
may be explored during clause learning. Note that these settings have
to be selected with care. Since several of the learned
programs employ recursion, a too restrictive proof depth can cause the
learner to incorrectly conclude that a candidate program has failed to
cover a training example. This would result in the addition of at best
redundant, and at worst harmful clauses. A bound on the proof depth is
nevertheless a necessity, as not all candidate programs are guaranteed
to terminate.

\subsection{Results}
\begin{figure}
  \centering
  \includegraphics[width=0.65\linewidth]{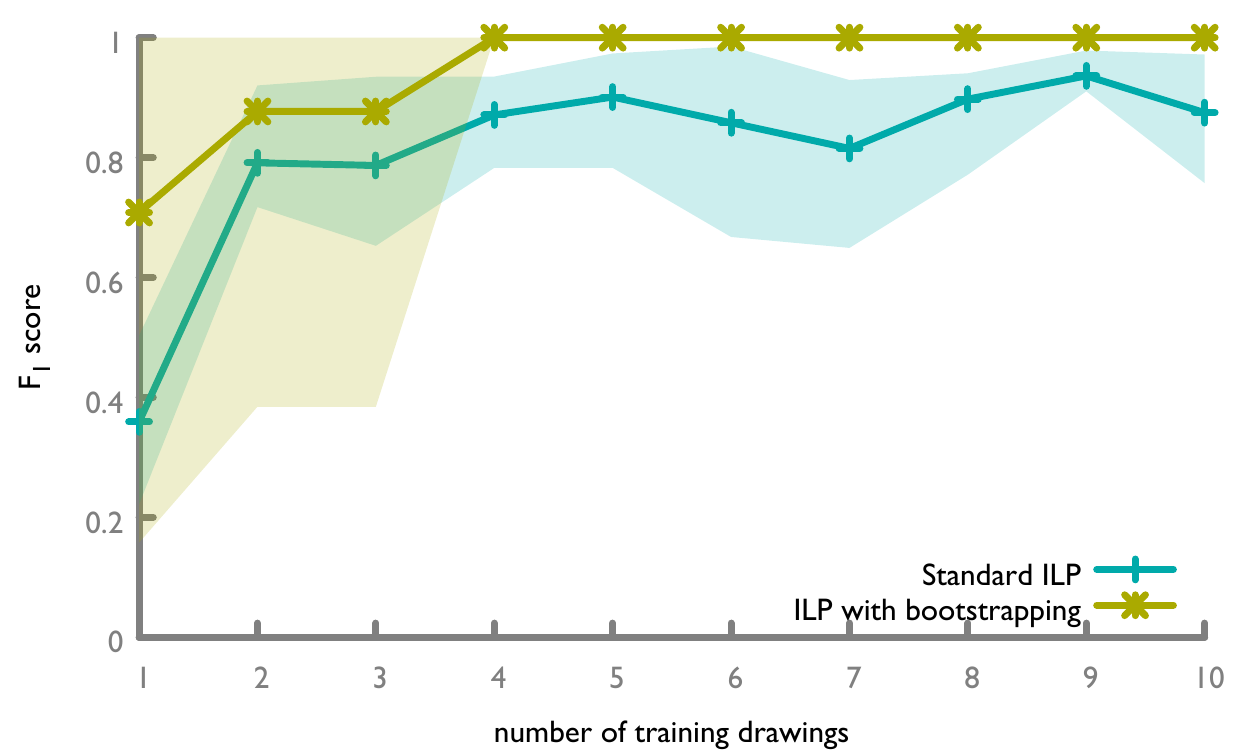}
  \caption{The performance measured using the $F_{1}$ score of programs learning materials/2. Min/max shading is included to indicate the range of performance between the best and worst-performing program over 5 repetitions.}
  \label{fig:f1-score-uncurated}
\end{figure}

\begin{figure}
  \centering
  \includegraphics[width=0.65\linewidth,angle=270]{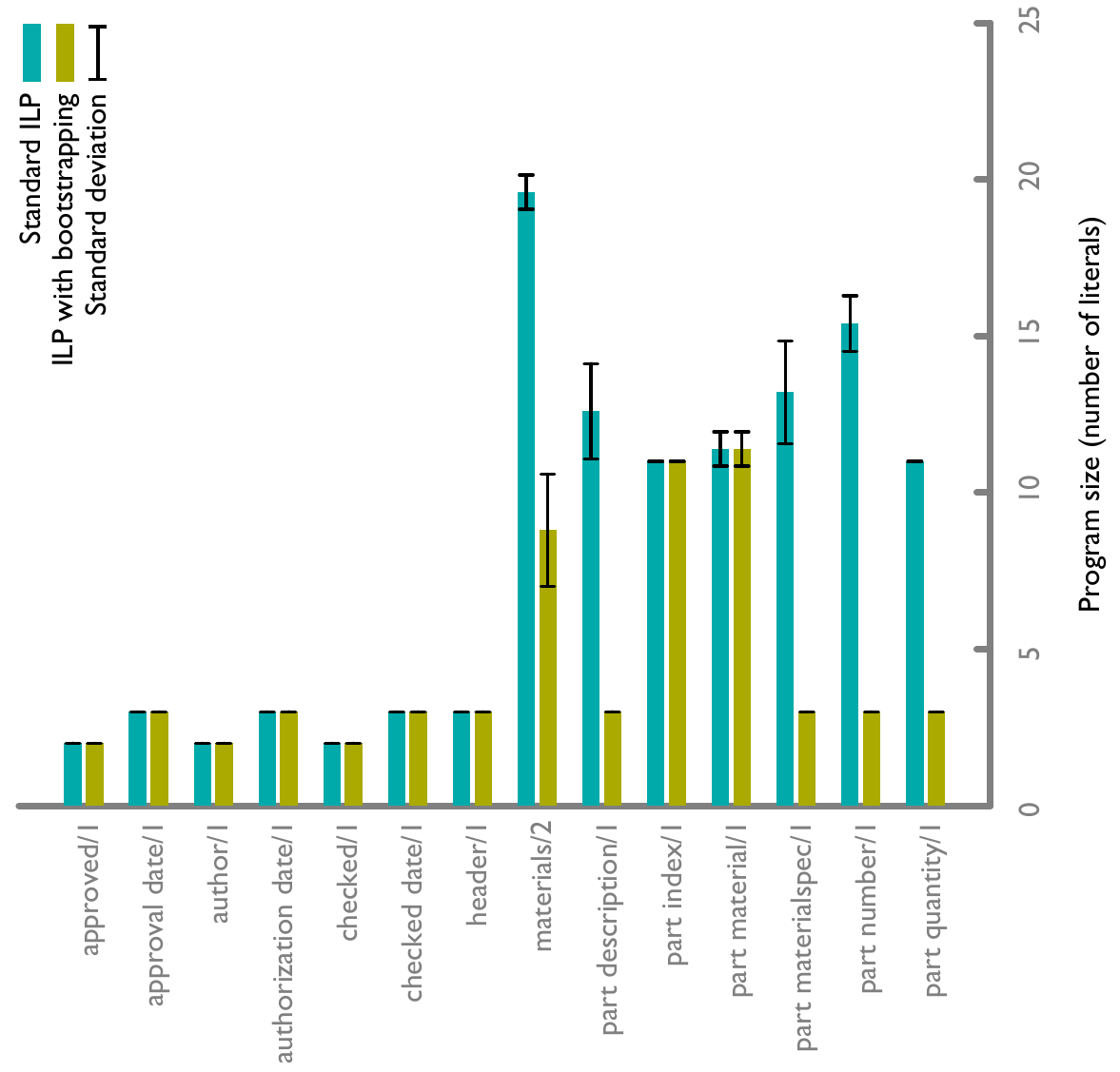}
  \caption{The difference in average program size (measured in number
    of literals) between standard ILP and
    ILP with bootstrapping across programs induced on 14 labels.
    10 repetitions were performed on a training set of 20 labeled
    drawings.
  }
  \label{fig:literal_count}
\end{figure}

Figure \ref{fig:f1-score-uncurated} visualizes the performance with
which cell labels and their appropriate index are correctly
identified. This shows that only a few annotated designs are required
for the bootstrapping method to learn perfect parsers for all labels
whereas the standard ILP approach fails to learn a perfect
parser. Furthermore, it highlights how ILP with bootstrapping compared
to Standard ILP is less sensitive to overfitting when presented with
additional training data. This robustness of ILP with bootstrapping
lends itself well to incremental learning. Both subtle and drastic
variations in template design can be handled by providing the learner
with a representative sample as training data.

We have observed that learning perfect parsers for simple labels such
as author or approval date can be achieved by both standard ILP and
the bootstrap method with only a few training examples. More
interesting is to look at the most complicated label, the indexed
components (\emph{materials} in Figure \ref{fig:fig1}).

The best performing program constructed using standard ILP in Figure
\ref{fig:f1-score-uncurated} consists of 14 clauses and yields 17
false negatives. Bootstrap learning, however, succeeds at learning a
completely accurate, concise program whenever more than three
technical drawings are provided in the training set. The poor
performance when using only a few drawings is due to poor
generalization. More specific, in these drawings the materials tables
provided for training each consisted only of a single row in addition
to the header, and there was no pressure on the inductive learner to
learn the recursive rule necessary to capture the rows of larger
tables.

A comparison of program size between Standard ILP and ILP with
bootstrapping across all investigated labels is shown in Figure
\ref{fig:literal_count}.
\subsection{Statistical Relational Learning for Uncertain Input}
The insights gained while learning the structure of the technical drawing using ILP can be leveraged to improve the knowledge extraction process.
Here we examine two improvements where the output of the optical character recognition (OCR) software is improved by augmenting it with contextual knowledge.
An issue is that the knowledge learned by ILP is deterministic while the results from OCR are probabilistic (i.e. a distribution over possible characters). A natural generalization of ILP is Statistical Relational Learning which combines logic formulas with probabilities. We use ProbLog, a probabilistic logic programming language, to extend the learned ILP programs to be able to cope with probability distributions.

\subsubsection{Probabilistic Levenshtein distance}
The content of many cells in a technical drawing are drawn from a predictable set of characters, and in some instances even from a specific set of strings. The employees involved in authoring a document are for instance likely to be present in an existing database. By explicitly establishing such a link, we can not only improve the quality of the text recognition, but also identify diacritics that are lost when names are for example displayed using upper-case representations, and avoid, or more easily identify and resolve ambiguity. While the Levenshtein distance is a suitable metric for identifying the similarity of pairs of strings, it fails to account for the probabilistic nature of our data. As such, we consider a generalization implemented in ProbLog that applies the Levenshtein distance to a probabilistic input string (see Listing~\ref{listing:pld}). The algorithm is similar to the implementation of Levenshtein using the Viterbi decoding except that penalties are expressed as probabilities, i.e. the probability of a transition in the Viterbi lattice, and the input is represented as a probability distribution over possible characters on each position in the string.

\begin{lstlisting}[basicstyle=\footnotesize, caption={A ProbLog
    program that performs a probabilistic Levenshtein distance (note that selecting the max-product semiring in ProbLog for this program achieves a probabilistic implementation of the Viterbi decoding version of the Levenshtein algorithm).},label={listing:pld},
  mathescape]

% OCR probability distribution
0.8::$c_o$(0,d); 0.1::$c_o$(0,b); 0.1::$c_o$(0,o).
$c_o$(1,r). $c_o$(2,i).
0.8::$c_o$(3,e); 0.2::$c_o$(3,3).
$c_o$(4,s). $l_o$(5).

% Probabilistic Levenshtein Distance (pld)
pld(_,-1,-1).
0.3::pld(W,R,C) :- % Insertion
	R>-1,C>-1, Rp is R-1,
	pld(W,Rp,C).
0.3::pld(W,R,C) :- % Deletion
	R>-1,C>-1, Cp is C-1,
	pld(W,R,Cp).
pld(W,R,C) :-      % Match
	R>-1,C>-1, Rp is R-1, Cp is C-1,
	nth0(R, W, CW), $c_o$(C, CO), CW=CO,
	pld(W,Rp,Cp).
0.3::pld(W,R,C) :- % No match
	R>-1, C>-1, Rp is R-1, Cp is C-1,
	nth0(R, W, CW), $c_o$(C, CO), CW\=CO,
	pld(W,Rp,Cp).
pld(W) :-          % Entry point
	length(W, WL), $l_o$(OL),
	WLp is WL - 1, OLp is OL - 1,
	pld(W, WLp, OLp).

% Query
query(pld(wannes)).  % Pr = 0.05
query(pld(dii3s)).   % Pr = 0.26
query(pld(dries)).   % Pr = 0.77
\end{lstlisting}

\subsubsection{Type-information enhanced OCR}
As a second use case we examine how integrating type-information can augment text
recognition.
The concept is illustrated on a cell in the \emph{quantity} column
of the materials table. Such a cell indicates the frequency with which its associated part
occurs within a design.

In our tests, text recognition frequently confused characters like `0'
with  `O', and `1' with characters such as `]' and `I'.
Such ambiguity is generally resolved through the use of dictionaries,
or by training the OCR software against the exact font used in the
example image.
In the \emph{quantity} column however, characters occur in isolation, and
even in the rest of the table they frequently appear as part of
complex identification codes. This renders publicly available dictionaries
useless. Furthermore, the robustness of the system demands high
performance even when the employed font is unknown.

Our proposed solution involves taking into account our expectation of
the type of data contained in a given cell. There exists for example a high 
degree of confidence that the \emph{quantity} column contains numerical
information. Since we now have the means to learn a program that
identifies these elements, the remaining challenge involves deciding
on how to combine this information. \cite{chan2005revision}
present a solution using virtual evidence. The virtual evidence
method considers an original distribution over events $\alpha$ (The
distribution over characters based on our understanding of the type
information) which is put up for revision based on some uncertain
evidence $\eta$ on a set of mutually exclusive and exhaustive events
$\gamma$ (the distribution over the set of characters returned by the
OCR software). It finds that virtual evidence can be incorporated
using Bayes' conditioning.

$$ P(\alpha | \eta) = \frac{\sum_{i=1}^n P(\eta | \gamma_i) P(\alpha,
  \gamma_i)}{\sum_{j=1}^n P(\eta | \gamma_j) P(\gamma_j)}$$

\begin{figure}[hbt]
\begin{subfigure}{.2\textwidth}
  \includegraphics[width=.5\linewidth]{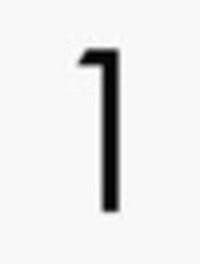}
  \caption{A cell containing `1'.}
  \label{fig:sub1}
\end{subfigure}%
\begin{subfigure}{.8\textwidth}
  \centering
  \begin{adjustbox}{max width=\textwidth}
        \begin{tabular}{cccc}
          symbol & OCR distribution & \multicolumn{2}{c}{OCR + type information}\\
          \hline
          {} & uncertain evidence & prior & posterior\\
          \hline
           1 & 0.130 & 0.615 & 0.544\\          
            ] & 0.630 & 0.077 & 0.330\\
          $\vert$ & 0.071 & 0.077 & 0.037\\
            I & 0.071 & 0.077 & 0.037\\
           J & 0.054 & 0.077 & 0.028\\
          l & 0.044 & 0.077 & 0.023\\
        \end{tabular}
        \end{adjustbox}
       \caption{Distributions over symbols for the identification of a single character}
  \label{fig:sub2}
\end{subfigure}
\caption{OCR is applied on the displayed cell. While the most likely
  candidate symbol according to the OCR distribution is the square
  bracket, the posterior correctly considers `1' to be the most
  probable symbol. Here, the prior distribution is defined only over
  those characters that appear in the OCR distribution.}
\label{fig:ocr}
\end{figure}

The set of events $\alpha$ has a one-to-one mapping with
$\gamma$ as they span the same set of possible characters. Since
$P(\alpha, \gamma_i) = P(\alpha | \gamma_i)P(\alpha)$, we can conclude that $P(\alpha | \gamma_i)$ corresponds to the
Kronecker-$\delta$ over $\alpha$ and $\gamma$. As such we find that
the numerator consists of the conjunction of the distribution returned
by the OCR software and the distribution established by the type
information, while the denominator acts as a normalizing term.
A ProbLog program  \citep{problog} representing virtual evidence applied to a particular cell of the
\emph{quantity} column is shown in Listing
\ref{listing:type-enhanced-ocr}, and its associated distributions are
explicitly shown in Figure \ref{fig:ocr}



\begin{lstlisting}[basicstyle=\footnotesize, caption={A ProbLog
    program that augments a given distribution $c_k$ using uncertain
    evidence $c_o$}\label{listing:type-enhanced-ocr},
  mathescape]

% For conciseness, we explicitly model $c_k$/1 as
% the prior distribution established by the
% type information (here, restricted to characters
% used by $c_o$).
% Normally, an implicit representation using rules
% would be employed
% e.g. [numerical:0.8, alphabetical:0.1, special:0.1]
0.615::$c_k$(1) ; 0.077::$c_k$(2) ; ... ; 0.077::$c_k$(]).

% OCR probability distribution
0.630::$c_o$(]) ; 0.130::$c_o$(1) ; 0.071::$c_o$(|) ;
0.071::$c_o$(I)  ; 0.054::$c_o$(J)  ; 0.043::$c_o$(l).

% Revising the distribution derived from
% the type-information by taking the conjunction
% with the OCR probability distribution
q(X) :- $c_k$(X), $c_o$(X).

% Setting the evidence
e :- q(X). % one of them should be true (i.e. 'there
% exists an answer that satisfies the query')
evidence(e).

% Query
query(q(X)).
\end{lstlisting}


\section{Extracting properties from CAD visualisation}
\label{sec:CNN}
Some aspects of a design can only be
communicated by sharing a visual depiction. This concerns for example
the subtleties of component shapes, and the exact manner of their
assembly. These features can prove essential in distinguishing
designs that have identical tabular information (e.g., material choices), and as such can be vital to construct a comprehensive representation of the technical drawing.

A technical drawing tends to contain both a 2D and 3D depiction. We
consider the 2-dimensional CAD drawing as the most suitable target for
visual analysis as the profile view offers a clear, uncluttered view
of the design.

\subsection{Learning key identifiers from unlabeled data}
\label{sec:learning-key-identifiers}
Many of these visual features of interest are only present in the CAD drawing and cannot be linked to features in the table or in the meta-data.
This means we cannot apply standard supervised learning because there are no labels available.
However, this is not a problem since we are mainly interested in learning what the relevant features are that can identify designs. Not yet what type of design it is.
The goal will thus be to learn a limited set of features that are expressive enough to uniquely identify designs, can generalize over different designs, and remain unaffected by translations or rotations of the design.

We propose to transform the problem to a binary classification task that captures these requirements. Given a pair of 2-dimensional CAD drawings, a classifier with a limitation on the numbers of features is trained to predict whether the pair represents the same design or not.
If the classifier achieves high accuracy on this task we consider the set of learned features to fit the requirements and we will use this set of features in a next step to summarize each design.

The data set for these input pairs is constructed as follows. For each
of the 2-dimensional CAD drawings available to us, we generate 10 variations by applying arbitrary flips, rotations and translations. We consider this image set consisting of 11 images to be representative for each design. The data for the `same' class is then formed by considering every pairing within each image set, while the `different' class is constructed by sampling an equal number of pairs across image sets. This ensures the resulting data set is balanced.

\subsection{CNN architecture and setup}
\label{sec:CNN-architecture}

Neural networks are widely known for their capability of capturing complex and non-obvious properties of the data they are trained with.
Convolutional Neural Networks (CNNs) are a category of neural networks of particular interest, as they have seen wide adoption in image recognition and classification. They are classifiers whose key characteristic is their usage of convolutional layers. An input image is passed through a series of such layers. Each layer consists of convolved features (i.e., the neurons) by applying a kernel on parts of its input. While the features captured in early layers are limited to angled edges and simple blobs, they become increasingly more complex as the layers deepen, until they are capable of describing domain-specific elements. 

\begin{figure}[h]
  \centering
  \includegraphics[width=0.8\linewidth,angle=0]{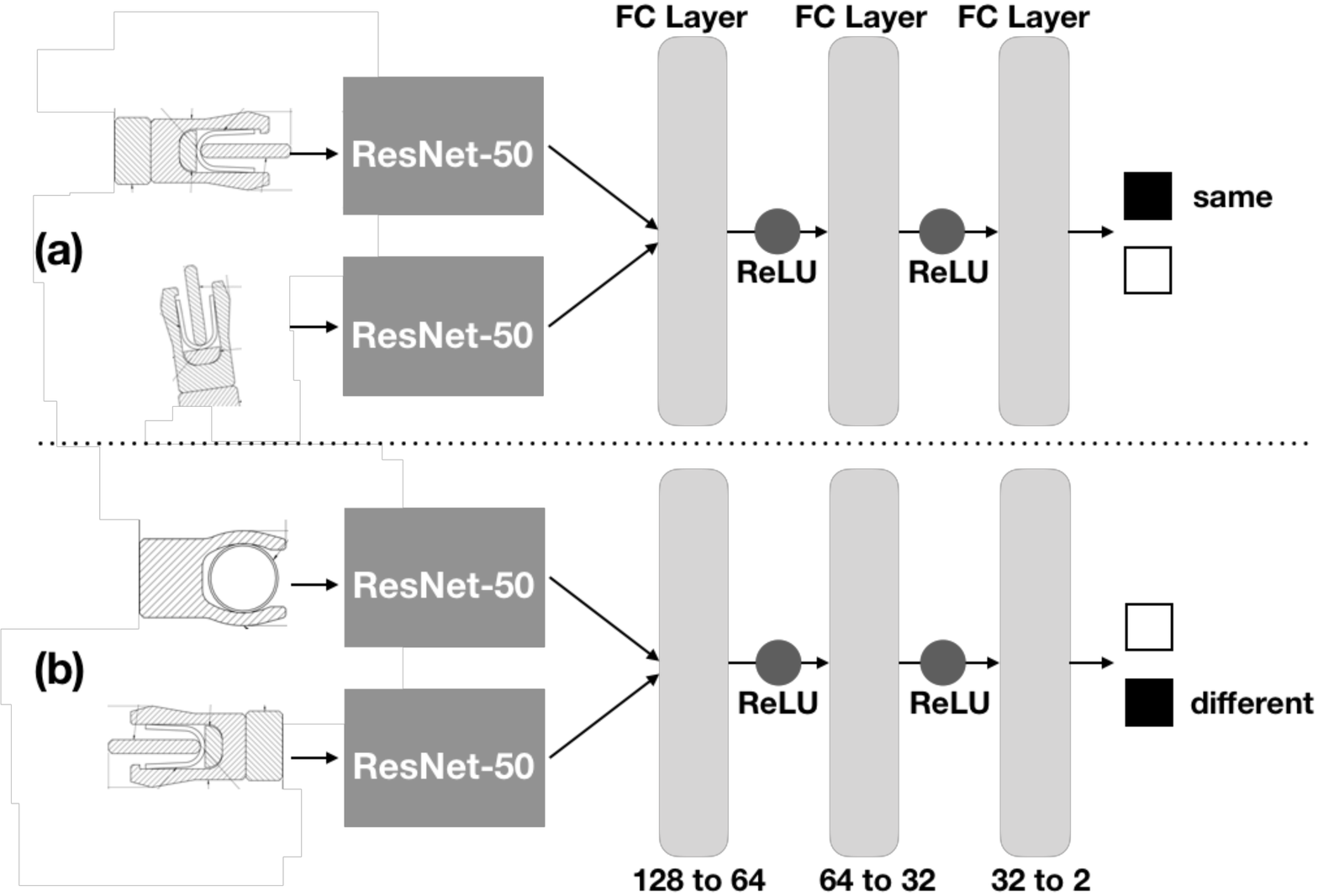}
  \caption{The ResNet-50 block consists of a ResNet-50 architecture
    where its final fully connected (FC) layer maps its 2048 inputs to
    a layer of size 64. This layer represents the set of derived features,
    i.e., the visual features of interest.}
  \label{fig:CNN_architecture}
\end{figure}

Convolutional neural networks have a proven capacity to learn complex features (i.e., representation learning). The challenge is to identify those that match our requirements. Notably, we find that the classification task outlined in Section \ref{sec:learning-key-identifiers} can directly be integrated in a CNN architecture, allowing it to learn features that are optimized to score well on the classification task.

Figure \ref{fig:CNN_architecture} shows our CNN
architecture. The ResNet-50 \citep{DBLP:journals/corr/HeZRS15} architecture is used to
perform the various convolutions.  Since the features identified in
the pre-trained model are tailored to a wide set of common settings,
while ours is a very domain-specific, we re-train only the last layer to
tailor the derived features to our data. While the output layer of
ResNet-50 has size 1000, we map it to a layer of size 64. This layer
corresponds to the relevant features that will be used to identify design. The rest of the architecture is
constructed in order to effectively classify the input image pairs
into one of the two possible classes: `same', or `different'.
Note that the same ResNet-50 network is used to encode both
images. The expectation is that the CNN to be attentive of visual
features of varying detail, as the visual differences between seals
can vary from striking to intensely subtle. Our classifier achieves a
96.8\% accuracy at this task (training
set: 68,507 pairs, validation set: 34,253 pairs, test set: 68,507 pairs).

\subsection{Experimental results}
\label{sec:CNN-results}

\subsubsection{t-Distributed Stochastic Neighbor Embedding (t-SNE)}

\begin{figure}[!htb]
 \centering
\begin{subfigure}{0.32\textwidth}
  \includegraphics[width=\linewidth]{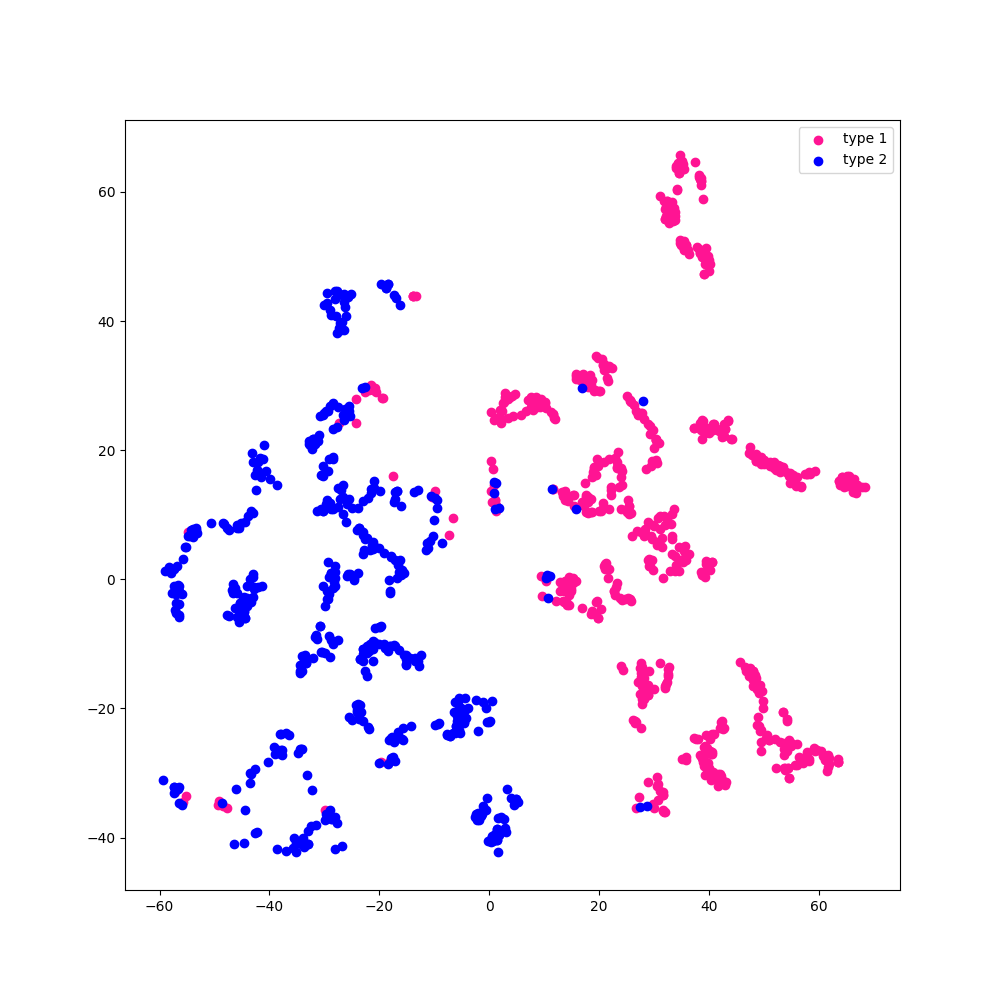}
  \caption{perplexity 20}\label{fig:t-SNE-20}
\end{subfigure}\hfill
\begin{subfigure}{0.32\textwidth}
  \includegraphics[width=\linewidth]{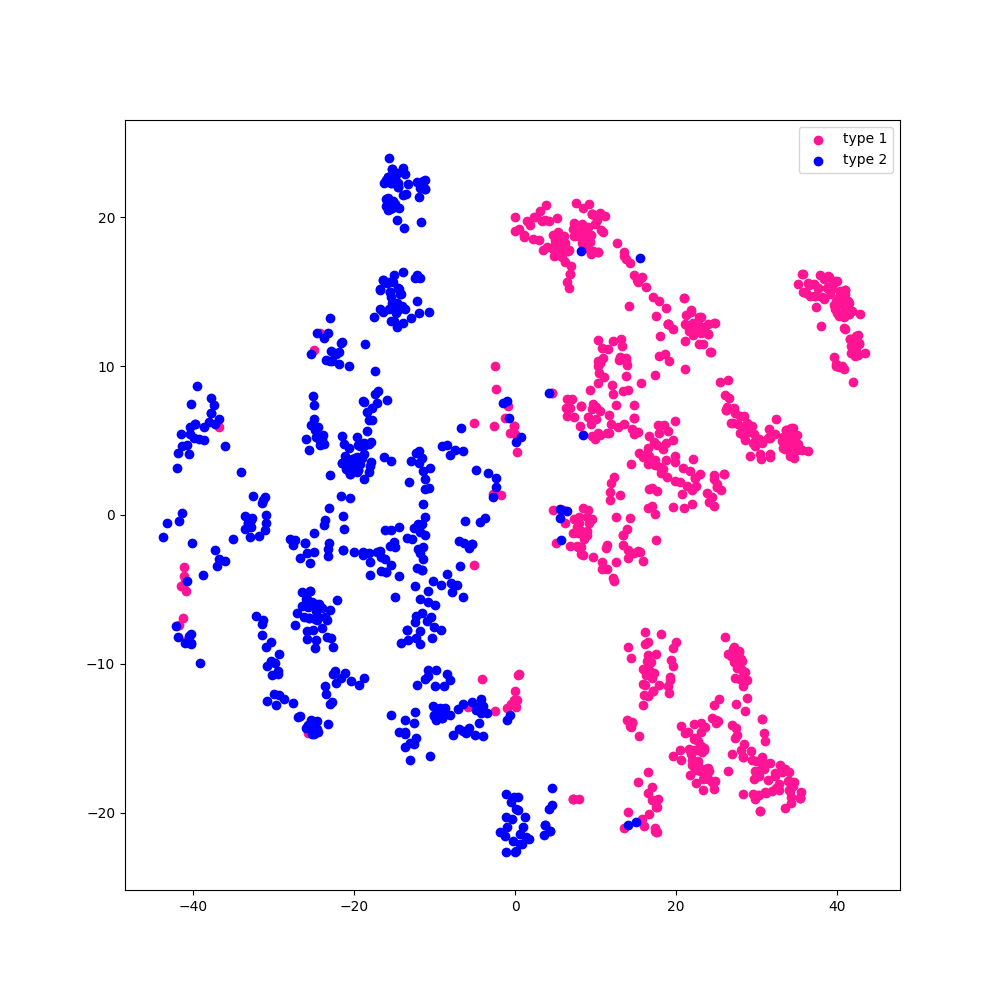}
  \caption{perplexity 50}\label{fig:t-SNE-50}
\end{subfigure}\hfill
\begin{subfigure}{0.32\textwidth}%
  \includegraphics[width=\linewidth]{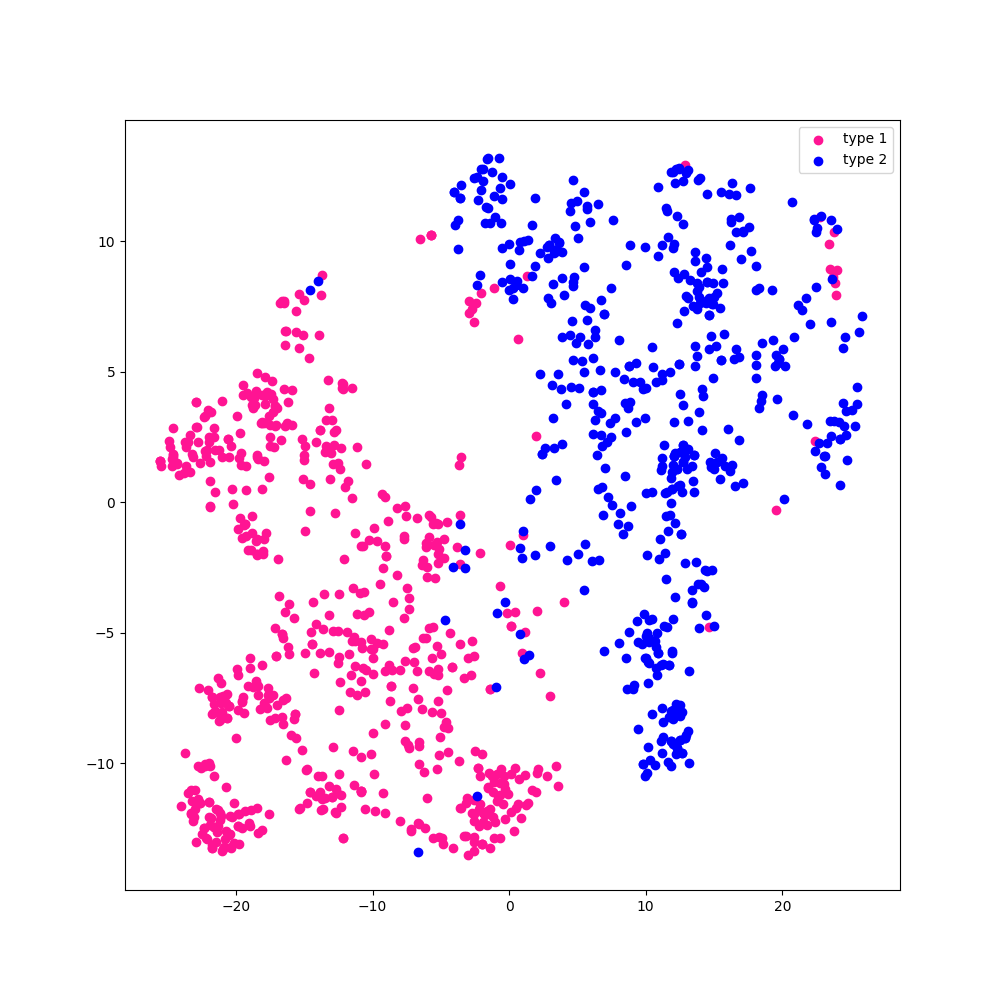}
  \caption{perplexity 100}\label{fig:t-SNE-100}
\end{subfigure}
\caption{t-SNE visualisation over 2 components for varying values of perplexity.}
\label{fig:t-SNE}
\end{figure}

We can now apply the CNN network to all drawings in the database and construct a feature vector from the 64 relevant features, thus the neurons in the last layer of our ResNet-50 model.
Figure \ref{fig:t-SNE} shows a t-SNE visualisation for varying values
of perplexity of the features vectors for each of the
drawings in our dataset. The data points are colored according to an
expert labeling which groups designs according to properties deemed to
have a high visual impact. This labeling has two values with high representation, and the visualisation clearly separates them in distinct, non-overlapping clusters. Since our interest lies in the detection of novel features, being able to identify a pre-existing one is not actually our goal. However, since a failure to visually distinguish between these types would falsify our hypothesis that we are extracting meaningful visual data, this result does inspire confidence.

\subsubsection{Gradient-weighted Class Activation Mapping (Grad-CAM)}
\begin{figure}
 \centering
  \includegraphics[width=0.8\linewidth]{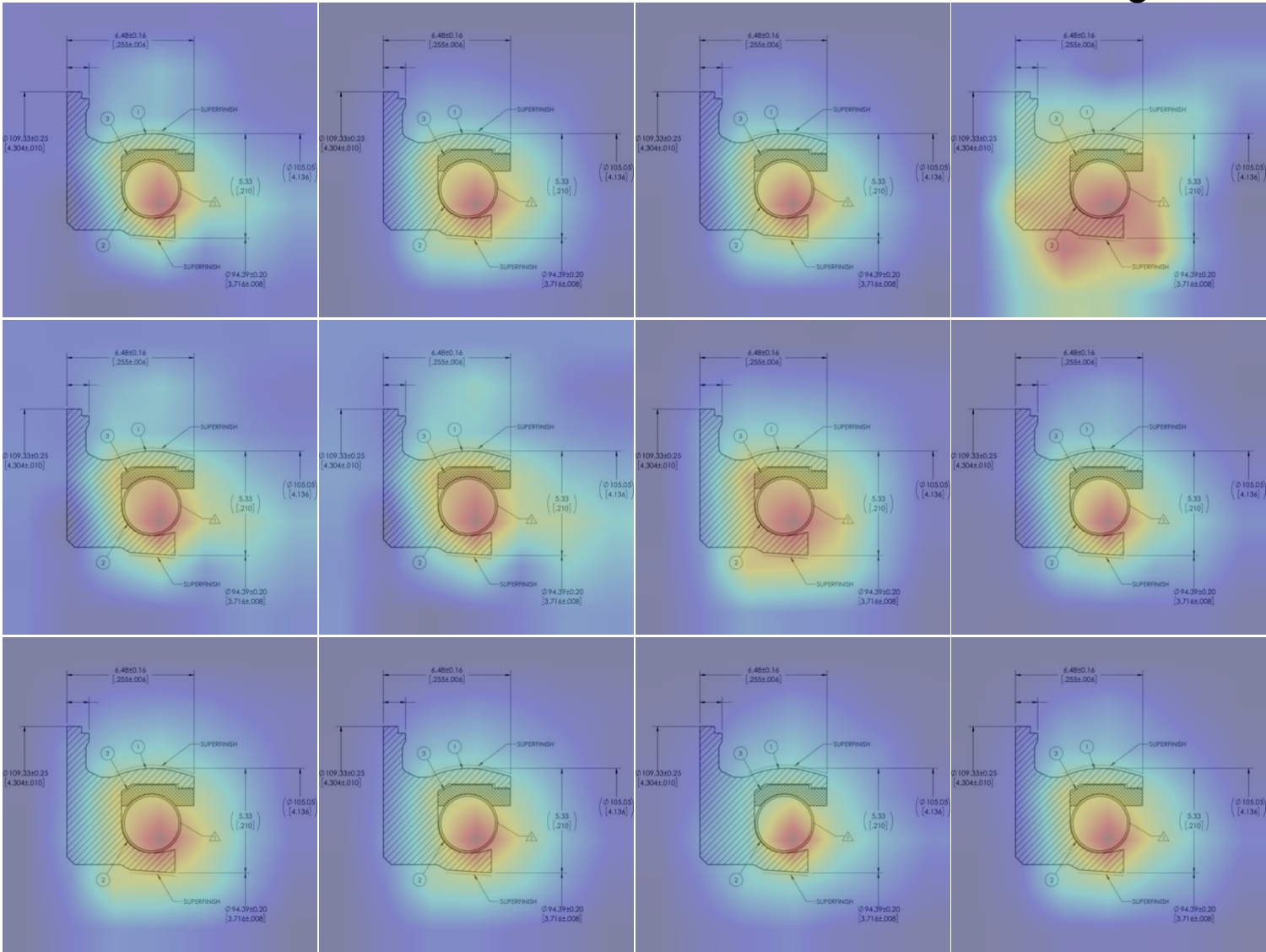}
  \caption{A Grad-CAM visualization. Warmer areas correspond to
    regions that play a more significant role in the activation of a
    derived feature. Here, each image depicts a derived feature whose attention seems largely focused on the ring of the seal}\label{fig:grad-cam-ring}
\end{figure}

This belief is strengthened further when analyzing the behaviour of
the CNN using Grad-CAM
\citep{DBLP:journals/corr/SelvarajuDVCPB16}. Grad-CAM provides a coarse
localization map of the important regions in the image. We find that
our derived features focus their attention on image regions that
correspond to meaningful properties in our application domain. Figure
\ref{fig:grad-cam-ring} visualizes a selection of the derived features activating in the presence of a particular type of ring seal.

\hfill


\section{Identifying relevant designs}
\label{sec:similarity-measure}
Using our results from Section \ref{sec:ilp-and-srl} and
\ref{sec:CNN}, we now have access to the design properties contained
in both the tabular data and the visual depiction of the seals. In
order to leverage these results when suggesting relevant designs, we
will first impose a similarity measure on them separately, and then
combine the resulting similarity measures using a weighted geometric mean.

\subsection{Tabular data similarity}
\label{sec:tabular-data-similarity}
Due to its highly relational nature, it is not feasible to directly
apply a conventional similarity measure to the tabular data. Instead,
we first propositionalize the data using Frequent pattern (FP) mining.
FP mining has shown to be effective at this task \cite{kramer2001propositionalization}. 

\begin{lstlisting}[basicstyle=\footnotesize, caption={Three frequent
    patterns mined using WARMR}\label{fig:frequent-pattern-examples}]
% A design has a part that uses the words
% `Double' and `NICKEL' in its cells.
materials(B,A),cell_contains(A,`Double'),
materials(B,C),cell_contains(C,`NICKEL').

% A design has a part that includes
% the word `COBALT' in one of its cells,
% and also employs the word `Spring'
% in its description column.
materials(B,A),cell_contains(A,`COBALT'),
materials(B,C),part_description(C),
cell_contains(C,`Spring').

% A design contains the words `Jacket',
% `Spacer', and `Spring' in its
% description column.
part_description(A),cell_contains(A,`Jacket'),
part_description(B),cell_contains(B,`Spacer'),
part_description(C),cell_contains(C,`Spring').
\end{lstlisting}

We perform FP mining using WARMR \citep{dehaspe1999discovery} on the
tabular data extracted in Section \ref{sec:ilp-and-srl}. We focus on
retrieving patterns that occur in $\ge 10\%$ of the drawings. Listing
\ref{fig:frequent-pattern-examples} shows a sampling of the resulting
patterns. In total we mined 9120 such patterns. Each technical drawing
is then represented as a binary vector indicating which patterns
are applicable. Given such a vector representation, closely related
designs can be identified by performing a ranking using the complement of its
normalized Hamming distance to all other designs.

Given two drawings represented as binary vectors X and Y
$$ \textit{sim}_{\textit{tabular}}(X, Y) =  1-\frac{1}{n}\sum_{i=1}^n| X_i - Y_i | $$

\subsection{Visual similarity of seals}
In Section \ref{sec:CNN} we discussed how the final fully connected
layer of our ResNet-50 architecture captures key visual properties of
a seal using a layer of size 64. All the features in this layer are
continuous. The expectation is that the features of visually similar
seals have similar values. Cosine similarity against this set of
derived features is used to determine similarity.

Given two drawings represented as feature vectors X and Y
$$ \textit{sim}_{\textit{visual}}(X, Y) = \frac{X \cdot Y}{||X||_2 \cdot ||Y||_2}$$

\subsection{Ranking designs by similarity to a given design}
\label{sec:ranking-similarity}




Tabular data similarity and visual similarity are combined using a
weighted geometric mean. 
$$\prod_{i}  \Bigl(x_i^{w_i} \Bigr)^{1/\sum_{i} w_i}$$
The use of a geometric mean ensures that
the resulting similarity is not biased towards a particular one of its
constituent similarity measures due to the distribution of their
values, 
but rather accounts only for the relative changes to each score across
designs. The use of a weighted mean makes the inherent trade-off
between potentially conflicting measures explicit.

\begin{figure}
  \centering
  \includegraphics[angle=0,width=0.8\linewidth]{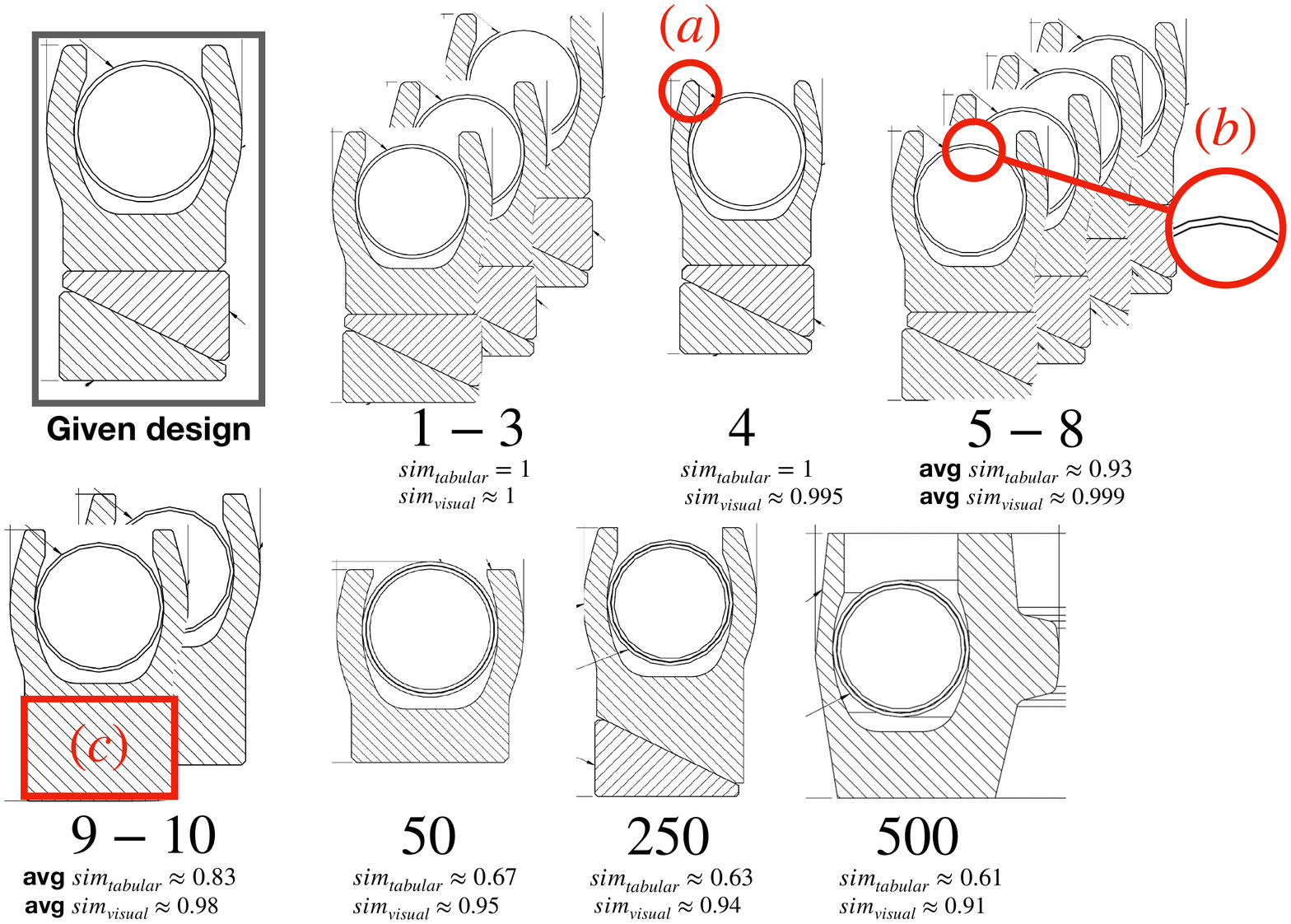}
  \caption{A depiction of a number of designs annotated with a rank
    indicating their similarity to the highlighted design
    ($\alpha=0.5$).
Subtle though significant differences are seemingly picked up, while insignificant though
sizable differences in shading seem to have little to no effect. The
lip of the seal in (a) differs 
from the original, possibly explaining the difference in its
$sim_{visual}$ score. The ring in (b) appears as a polygon instead of
a circle, and the body of (c) consists of a single contiguous part as
opposed to multiple elements separated by a diagonal.
  }
  \label{fig:similar_designs}
\end{figure}

Here, we use this convex combination solely to combine tabular data
similarity and visual similarity. This trade-off can be captured in a single parameter $\alpha$ such that%
$$\textit{similarity} = {(\textit{sim}_{\textit{tabular}})}^\alpha \cdot
{(\textit{sim}_{\textit{visual}})}^{1-\alpha}$$
In doing so, users can easily impose their own biases and
preferences to influence the ranking. In our implementation we use $\alpha=0.5$. Figure \ref{fig:similar_designs}
shows a depiction of some of the designs ranked given a particular
technical drawing.

\subsection{Ranking designs by similarity to a given partial design}
The ability to identify similar designs is particularly useful when
applied to a design under construction. If such a
design displays high similarity to existing ones, the most
similar ones can provide the engineer with possible completions.

We consider a partial design to be represented as a technical drawing
containing a number of empty cells. During property extraction, the
cell text of empty cells is represented using logical variables.

If a visual depiction of the seal is present, $sim_{visual}$ can be
computed as usual. If it is omitted, the similarity measure is
restricted to its $sim_{tabular}$ term.

As a consequence of allowing the inclusion of empty cells, some of the patterns employed by the feature vector used to
construct $sim_{tabular}$ might not have a proper instantiation on the
partial design. Given T the set of all properties of  the partial technical drawing, we
distinguish three situations:

{\setlength{\itemsep}{0pt}\setlength{\topsep}{0pt}\setlength{\partopsep}{0pt}\setlength{\parsep}{0pt}\setlength{\parskip}{0pt}%
\begin{itemize}
\item[--] \emph{True.} A feature is true on the partial design when each of the literals in its conjunction have a corresponding match in T that does not require the instantiation of any logical variable in T.
\item[--] \emph{False.} A feature is false on the partial design if at least one of the literals in its conjunction fails to find a corresponding match in T.
\item[--] \emph{Unknown.} Any feature for which it is not known whether
  it is \emph{True} or \emph{False}.
\end{itemize}
}

$sim_{tabular}$ against a given partial design can then be computed by
utilizing feature vectors that are filtered to contain only those features
that were \emph{True} or \emph{False} on the partial design.

\section{Related Work}
\subsection{Digitisation of technical drawings}

Two recent papers cover the domain of document digitisation. First, \cite{Staar_2018} shows significant progress digitising general PDF and bitmap documents. Second, \cite{Moreno-Garca2018} provides an overview of all recent trends on digitising engineering drawings in particular.
%
%
The work of \cite{Staar_2018} focuses on detecting elements in text documents in general and parses tables but cannot take easily into account expert knowledge to achieve near perfect extraction for specific cases or extract information embedded in figures.

With regards to shape
detection, \cite{Moreno-Garca2018} makes a distinction between \emph{specific} approaches
focused on identifying shapes that are known in advance, and
\emph{holistic} approaches where the underlying rules of the drawing
are exploited to split it into parts. Our approaches for identifying
technical drawing elements utilized a mixture of both. The use of
image segmentation and text detection in Section~\ref{sec:preprocessing} falls under the holistic view,
while the contour detection in Section
\ref{sec:object-recognition} is an illustration of a specific approach.
While \cite{Moreno-Garca2018} notes that some frameworks do perform contextualisation
 (i.e., inferring and exploiting the relationship between symbols within drawings), our approach is to our knowledge the first that
 enables the construction of a comprehensive, formal representation of
 a technical drawing that supports the inclusion of expert knowledge
 that is obtained by learning an interpretable parser that takes into
 account expert knowledge and can identify unique properties in drawings.

\subsection{Feature extraction}
While established feature extraction methods such as SIFT \citep{lowe2004distinctive} and SURF
\citep{bay2006surf} are still viable alternatives, CNNs are increasingly the go-to method when extracting features. 
Industrial applications of CNNs are however strongly limited by the
cost of collecting a suitable set of labeled training data
\citep{Moreno-Garca2018}. Our approach side-steps this issue by
learning from unlabeled data through the introduction of a discriminative setting. This approach is comparable to that of Exemplar-CNN \citep{NIPS2014_5548}.

In this work, autoencoders were observed to perform poorly as a
means to capture features. When encoding an input image to a
lower dimensional feature space, an autoencoder seeks to capture as
much of the input data as possible with the aim of later on
reconstructing the image as faithfully as possible.
Here however, most of the input data proves to be
irrelevant. The exact position and rotation of a seal is completely
irrelevant, and even though the shading of a seal represents a large
amount of data, it is not something that merits encoding. We found
that our proposed, discriminative approach is far more suitable for
identifying notable features.

\subsection{Inductive Logic Programming systems}
The ILP system Aleph was used for the parser learning discussed in Section \ref{sec:ILP-set-up}. Related are all the ILP systems that currently define the state-of-the-art. This includes Tilde \citep{blockeel1997experiments}, Aleph \citep{srinivasan2001aleph}, Metagol \citep{cropper2016learning}, Progol \citep{muggleton1995inverse}, and FOIL\citep{quinlan1993foil}.

While Aleph learns from entailment, Tilde learns a relational decision
tree from interpretations. Both systems were considered, but only
Aleph was capable of constructing recursive programs. This allows it
to construct concise programs, making Aleph our system of choice. FOIL is expected to be similarly suitable.
Metagol is expected to be highly effective at this task, as its metarules allow for a more targeted search for recursive programs. A downside of this system is that meta-rules are currently user-defined, imposing an additional burden on the user, who in this setting is a domain expert with no background in ILP. Automatic identification of metarules is ongoing work \citep{cropandr}.






\section{Conclusions and future work}

We introduced an approach to assist an engineer by automatically interpreting technical drawings and allowing for a flexible search method.
To achieve this we introduced five contributions.
First, we introduced the use of ILP to learn parsers from data and expert knowledge to interpret a technical drawing and produce a formal representation.
Second, we introduced a novel bootstrapping learning strategy for ILP.
Third, we introduced a deep learning architecture that learns a meaningful summarization of CAD drawings by identifying unique properties in drawings.
Fourth, we introduced a similarity measure to find related technical drawings in a large database.
Finally, the efficacy of this method was demonstrated in a number of experiments on a real-world data set.

From this work, more advanced techniques can be developed to achieve an automated engineering assistant. For example, given the interpreted technical drawings, one can learn constraints or rules that apply to a given set of designs. Such rules can then later be used to automatically verify novel designs or find anomalous designs by identifying constraints that are violated.

%

\section*{Acknowledgements}
The authors would like to thank Luc De Raedt for his many suggestions and critical feedback.
This research is supported by VLAIO-O\&O project ``Digital Engineer for Seals''.
This work has received funding from the European Research Council (ERC) under the European Union’s Horizon 2020 research and innovation programme (grant agreement No 694980, SYNTH: Synthesising Inductive Data Models).

\section*{References}
\bibliographystyle{ACM-Reference-Format}
\bibliography{sample-base}
\end{document}